\documentclass[10pt, a4paper, fleqn]{article}
\usepackage{lrec2006}
\usepackage{graphicx}
\usepackage{url}
\expandafter\def\expandafter\UrlBreaks\expandafter{\UrlBreaks
  \do\a\do\b\do\c\do\d\do\e\do\f\do\g\do\h\do\i\do\j%
  \do\k\do\l\do\m\do\n\do\o\do\p\do\q\do\r\do\s\do\t%
  \do\u\do\v\do\w\do\x\do\y\do\z\do\A\do\B\do\C\do\D%
  \do\E\do\F\do\G\do\H\do\I\do\J\do\K\do\L\do\M\do\N%
  \do\O\do\P\do\Q\do\R\do\S\do\T\do\U\do\V\do\W\do\X%
  \do\Y\do\Z}
  
\usepackage{hyperref}  
\usepackage{amsmath}
\newcommand\addtag{\refstepcounter{equation}\tag{\theequation}}

\usepackage{array}
\newcolumntype{P}[1]{>{\centering\arraybackslash}p{#1}}

\title{On TimeML-Compliant Temporal Expression Extraction in Turkish}

\name{Dilek K\"u\c{c}\"uk$^{\star}$, Do\u{g}an K\"u\c{c}\"uk$^{\ast}$}

\address{$^{\star}$T\"UB\.ITAK Energy Institute\\
Ankara{--}Turkey\\
dilek.kucuk@tubitak.gov.tr\\\\
$^{\ast}$Gazi University\\
Ankara{--}Turkey\\
dogan.kucuk@gazi.edu.tr\\}

\abstract{It is commonly acknowledged that temporal expression extractors are important components of larger natural language processing systems like information retrieval and question answering systems. Extraction and normalization of temporal expressions in Turkish has not been given attention so far except the extraction of some date and time expressions within the course of named entity recognition. As TimeML is the current standard of temporal expression and event annotation in natural language texts, in this paper, we present an analysis of temporal expressions in Turkish based on the related TimeML classification (i.e., date, time, duration, and set expressions). We have created a lexicon for Turkish temporal expressions and devised considerably wide-coverage patterns using the lexical classes as the building blocks. We believe that the proposed patterns, together with convenient normalization rules, can be readily used by prospective temporal expression extraction tools for Turkish. \\ \newline }

\begin{document}

\maketitleabstract

\section{Introduction}\label{sec:intro}
Temporal expressions in natural language texts stand as one of the crucial pieces of information to be extracted from these texts. Accordingly, several text analysis applications, like event extractors \cite{Ritter2012} and text-based video annotation systems \cite{Kucuk2011}, include a temporal expression extractor as a submodule to identify, normalize and then make use of these expressions.\\

Traditionally, temporal expressions like some date and time expressions have been considered as named entities and have been included in the scope of named entity recognition (NER) systems. For instance, in the Message Understanding Conference (MUC) series \cite{Grishman1996}, which have been conducted for several years to promote research in information extraction, some date and time expressions were considered within the scope of the NER task and, in the related guidelines, these expressions are recommended to be annotated with the \texttt{TIMEX} tag. But within the scope of MUC, only the identification of these temporal expressions was required without the need for their normalization.\\

TimeML is a standard markup language for annotating temporal expressions and events \cite{Pustejovsky2003} which is built upon previous work on the annotation of temporal expressions such as \cite{ferro2001tides,setzer2001temporal}. According to the current TimeML guideline \cite{saurii2005timeml}, \texttt{TIMEX3} tag is used to annotate the temporal expressions identified and the normalized forms of the expressions are also specified within the annotations. Additionally, \texttt{SIGNAL} tag is used to annotate the temporal relations between two temporal expressions, two events, or a temporal expression and an event. There are mainly four distinct temporal expressions within the scope of TimeML: date, time, set, and duration \cite{saurii2005timeml}. Hence, the extent of the temporal expressions considered in TimeML is also broader compared to the extent considered in the MUC series, in addition to the normalization procedure introduced.\\

There are several temporal expression extraction and normalization systems, as reported in studies like \cite{Uzzaman2013Semeval}. One of the initial such systems is called GUTime which is the temporal expression recognition and normalization module of a larger system called TARSQI which annotates temporal expressions, relations, and events in news texts \cite{Verhagen2005}. Several of the system proposals so far, including Edinburgh-LTG \cite{Grover2010Edinburgh}, HeidelTime \cite{HeidelTime}, SUTime \cite{Chang2013SUTime}, and FSS-TimEx \cite{FSS-TimEx} are rule-based systems and some of them, such as HeidelTime, have been extended to extract temporal expressions in other languages including Arabic, Italian, Spanish, and Vietnamese \cite{strotgen2014time}. As previously pointed out, the extraction of some temporal expressions has long been considered a subtask of named entity recognition, and accordingly, some of the aforementioned systems like Edinburgh-LTG and SUTime are based on previous NER systems.\\

In addition to the system proposals, as there is a need for corpora annotated with temporal expressions, relations, and events, resources like TimeBank \cite{timebank} have emerged. TimeBank has been commonly used to evaluate and compare different system proposals. Similar annotated resources have also been constructed for other languages, such as French TimeBank \cite{frenchtimebank}, Spanish TimeBank \cite{spanishtimebank}, and Italian Timebank \cite{italiantimebank}. Such resources are indispensable for training the extraction systems proposed in addition to the common evaluation and thereby comparison of different system proposals, and to the best of our knowledge, no such annotated resource exists for Turkish.\\

Considering the related tools on Turkish, extraction of date and time expressions has been performed by the rule-based NER system \cite{Kucuk2009} and its extended versions like \cite{Kucuk2012}, mostly following the named entity definition of the MUC series and extracting some deictic date expressions as well, without normalization. These experiments have been performed on diverse text genres such as news articles, historical texts, and child stories. Within a text-based semantic video annotation system, which makes use of this NER system, a separate date normalization module has been implemented to normalize only the deictic date expressions using the creation dates of the corresponding videos as reference dates \cite{Kucuk2011}. Within the course of this latter study, extraction experiments are performed on automatically obtained news video texts which are mostly noisy (due to the character recognition errors introduced during the sliding text recognition procedure employed). Recently, date and time expressions are also recognized in informal texts (i.e., tweets) in Turkish using the aforementioned rule-based system, as described in \cite{Kucuk2014_2}. Another related work is presented in \cite{Seker2010timeml}, where the authors have considered temporal logic and event times in Turkish based on existing temporal models, yet, it does not aim to propose a temporal expression extractor or related resource for Turkish.\\

In this paper, we provide an analysis of the temporal expressions in Turkish, following the corresponding TimeML classification. We mainly provide several wide-coverage patterns for the extraction of these expressions together with sample expressions and their annotated forms with the \texttt{TIMEX3} tag. With the presented lexicon, pattern bases, and the review of the related limited literature on Turkish, we believe that this paper can be used as a guideline before building a temporal expression extraction and normalization system for Turkish. The rest of the paper is organized as follows: In Section 2, a compact temporal lexicon in Turkish and patterns for the extraction of temporal expressions in Turkish are presented together with several samples. Section 3 lists the open issues on temporal expression extraction and normalization in Turkish texts and Section 4 concludes the paper.

\section{Temporal Expressions in Turkish}\label{sec:tempex}
Before presenting the lexical resources and patterns for temporal expressions in Turkish, we briefly summarize their two particularities in formal Turkish texts which should also be considered during system development. These writing rules are provided below, following the corresponding language rules published by \emph{T\"urk Dil Kurumu} (`\emph{Turkish Language Association}') \cite{YazimKilavuzu2015}:
\begin{itemize}
    \item   The tokens within temporal expressions are all in lowercase, except the names of the months and week days which have their initial letters capitalized. Sample expressions are \emph{bug\"un} (`\emph{today}'), \emph{yar{\i}ndan sonraki g\"un} (`\emph{the day after tomorrow}'), \emph{Pazartesi sabah{\i}} (`\emph{Monday morning}'), \emph{May{\i}s ay{\i}n{\i}n ikinci Pazar g\"un\"u} (`\emph{the second Sunday of (the month of) May}').
    \item   The suffixes attached at the ends of the tokens of the temporal expressions are not separated from the attached suffixes. The names of the months and week days and numerals constitute the exceptions of this characteristic, as the sequence of suffixes added to the ends of these are separated from them with apostrophes. In the illustrative temporal expression, \emph{2015 y{\i}l{\i}n{\i}n Mart'{\i}n{\i}n 23'\"u} (`\emph{the 23rd of March of (the year of) 2015}'), the sequence of suffixes attached at the end of \emph{y{\i}l} (`\emph{year}') is not separated from it while the ones attached at the ends of the numeral (\emph{23}) and \emph{Mart} (`\emph{March}') are separated with apostrophes.
\end{itemize}

As mentioned in Section 1, there are four distinct types of temporal expressions within the scope of TimeML: date, time, set, and duration, which correspond to the value range of the \texttt{type} attribute of the \texttt{TIMEX3} tag. In this section, we first provide a compact temporal lexicon for Turkish and in the following subsections, we present patterns for temporal expressions in Turkish and then samples conforming to these patterns.\\

We should note that both the lexicon and the pattern bases are nowhere near exhaustive. We have tried to devise patterns with high coverage as much as we can, yet, they are all open to modifications, corrections, and extensions especially when building practical systems for Turkish. Normalization is also not considered within the current study, a distinct set of normalization rules should be devised for the extracted temporal expressions as part of the future work.

\subsection{Turkish Lexicon for Temporal Expressions}
We have built the Turkish lexicon for temporal expressions with the following lexical classes. The class identifiers are given in parentheses and they are used in the ultimate extraction patterns as the building blocks.
\begin{enumerate}
    \item   The list of cardinal numerals from 1 to 2100, both in numbers and in words (\texttt{<NUM>}), and the list of the corresponding ordinal numbers (\texttt{<ORD>}).
    \item   The names of days (\texttt{<DAY>}), that of months (\texttt{<MON>}), that of seasons (\texttt{<SEAS>}).
    \item   The names of the parts of a day, like \emph{sabah} (`\emph{morning}'), \emph{ak\c{s}am} (`\emph{evening}') etc. (\texttt{<D-PART>}).
    \item   The names of the units of time, like \emph{saat} (`\emph{hour}'), \emph{g\"un} (`\emph{day}') etc. (\texttt{<T-UNIT>}).
    \item   The modifiers of temporal expressions, like \emph{gelecek} (`\emph{next}'), \emph{ge\c{c}en} (`\emph{last}') etc. (\texttt{<MOD>}).
    \item   Deictic temporal expressions like \emph{\c{s}imdi} (`\emph{now}'), \emph{d\"un} (`\emph{yesterday}') etc. (\texttt{<DEIC>}).
    \item   The determiners like \emph{her} (`\emph{every}') (\texttt{<DET>}).
    \item   The quantifiers like \emph{kere} (`\emph{times}' as in \emph{three times a day}) (\texttt{<QUANT>}).
    \item   The suffixes that can be attached at the ends of temporal expressions like the case (including genitive and possessive) markers, plural markers, and relativizers in Turkish (a single such suffix is denoted as \texttt{<SUF>}).
    \item   The apostrophe character (\texttt{<APST>}).
\end{enumerate}

\subsection{Date Expressions}
Before presenting the actual patterns for date expressions (\texttt{<DATE-EXPR>}), we first present patterns for auxiliary constructs of \texttt{<DAY-EXPR>}, \texttt{<MON-EXPR>}, and \texttt{<YEAR-EXPR>} which are in turn used within the \texttt{<DATE-EXPR>} patterns. The patterns are presented as regular expressions where \texttt{?} denotes zero or one, \texttt{*} denotes zero or more, \texttt{$|$} denotes the OR operator and parentheses are for grouping purposes. The patterns may include both the classes of lexical entries, described in the previous section, and rarely individual entries themselves, like \emph{y{\i}l} (`\emph{year}'), \emph{sene} (`\emph{year}'), \emph{ay} (`\emph{month}'), \emph{g\"un} (`\emph{day}'), and \emph{saat} (`\emph{hour}').\\

Though not denoted in the patterns, there are also constraints, regarding the lexical entries, that should be enforced during the utilization of the patterns. For instance, the \texttt{<NUM>} values within the \texttt{<DAY-EXPR>} should be within the range of [1..31] while the \texttt{<NUM>} values within the \texttt{<MON-EXPR>} should be within the range of [1..12].\\

\noindent
\[
\texttt{<DAY-EXPR> $\rightarrow$ (<NUM><APST> | (<ORD> | <DAY>) g\"un)<SUF>*}
\]

\noindent
\[
\texttt{<MON-EXPR> $\rightarrow$ <MON><APST><SUF>* | (<ORD> | <MON>) ay)<SUF>*}
\]

\noindent
\[
\texttt{<YEAR-EXPR> $\rightarrow$ <NUM> ((y{\i}l | sene)<SUF>*)?}
\]

\vspace{5pt}

Below provided are some wide-coverage patterns for extracting date expressions in Turkish.\\

\noindent
\[
\texttt{<DATE-EXPR> $\rightarrow$ (<NUM>.<NUM>.<NUM> | <NUM>/<NUM>/<NUM>)}    \addtag
\]
\[
\texttt{<DATE-EXPR> $\rightarrow$ <NUM>? <MON> <NUM>? <DAY>?}    \addtag
\]
\[
\texttt{<DATE-EXPR> $\rightarrow$ <YEAR-EXPR> <MON-EXPR>? <DAY-EXPR>?}    \addtag
\]
\[
\texttt{<DATE-EXPR> $\rightarrow$ <YEAR-EXPR> <NUM> (<MON><SUF>* | <MON> <DAY>?)}    \addtag
\]
\[
\texttt{<DATE-EXPR> $\rightarrow$ <MON-EXPR> <DAY-EXPR>?}    \addtag
\]
\[
\texttt{<DATE-EXPR> $\rightarrow$ <MOD>? (<T-UNIT> | <DAY> | <MON> | <SEAS>)}    \addtag
\]
\[
\texttt{<DATE-EXPR> $\rightarrow$ <DEIC>}    \addtag
\]

\vspace{5pt}

Sample date instances conforming to some of these patterns are given in Table \ref{tab:dates}. In this table and the other tables in the current paper, the first column shows the Turkish samples, the second column shows their meanings in English, the third column shows the \texttt{TIMEX3} annotation of the sample, and the fourth column shows the number of the pattern that the sample conforms to. For the sample in the second to last row of Table \ref{tab:dates}, the normalized value is given with respect to a reference date in the year 2015.

\begin{table}[!h]
  \caption{Sample \emph{Date} Expressions in Turkish.}
  \label{tab:dates}
    \centering
    \begin{tabular}{p{0.24\linewidth}p{0.21\linewidth}p{0.37\linewidth}P{0.07\linewidth}}
    \hline
    \emph{Date Expression}&\emph{Meaning}&\emph{TIMEX3 Annotation}&\emph{Pattern}\\
    \hline
    23.03.2015&23.03.2015&\texttt{<TIMEX3 tid="t1" type="DATE" value="2015-03-23">23.03.2015 </TIMEX3>}&(1)\\
    \hline
    23 Mart 2015&March 23, 2015&\texttt{<TIMEX3 tid="t1" type="DATE" value="2015-03-23">23 Mart 2015</TIMEX3>}&(2)\\
    \hline
    23 Mart 2015 Pazartesi&March 23, 2015 Monday&\texttt{<TIMEX3 tid="t1" type="DATE" value="2015-03-23">23 Mart 2015 Pazartesi</TIMEX3>}&(2)\\
    \hline
    2015 y{\i}l{\i}n{\i}n Mart'{\i}n{\i}n 23'\"u&the 23rd of the March of the year 2015&\texttt{<TIMEX3 tid="t1" type="DATE" value="2015-03-23">2015 y{\i}l{\i}n{\i}n Mart'{\i}n{\i}n 23'\"u</TIMEX3>}&(3)\\
    \hline
    2015 y{\i}l{\i} 23 Mart'{\i}&the 23rd of the March of the year 2015&\texttt{<TIMEX3 tid="t1" type="DATE" value="2015-03-23">2015 y{\i}l{\i} 23 Mart'{\i}</TIMEX3>}&(4)\\
    \hline
    Mart ay{\i}n{\i}n ikinci g\"un\"u&the second of March&\texttt{<TIMEX3 tid="t1" type="DATE" value="XXXX-03-02">Mart ay{\i}n{\i}n ikisi</TIMEX3>}&(5)\\
    \hline
    ge\c{c}en sonbahar&last autumn&\texttt{<TIMEX3 tid="t1" type="DATE" value="2014-FA">ge\c{c}en sonbahar</TIMEX3>}&(6)\\
    \hline
    \c{s}imdi&now&\texttt{<TIMEX3 tid="t1" type="DATE" value="PRESENT\_REF">\c{s}imdi</TIMEX3>}&(7)\\
    \hline
    \end{tabular}
\end{table}

\subsection{Time Expressions}
Below listed are the patterns for the common time expressions in Turkish and samples conforming to these patterns are provided in Table \ref{tab:times}. As the final pattern denotes, some time patterns make use of date expressions extracted as well and can be recursive.\\
\noindent
\[
\texttt{<TIME-EXPR> $\rightarrow$ <D-PART>? saat? (<NUM>.<NUM> | <NUM>:<NUM>)}    \addtag
\]
\[
\texttt{<TIME-EXPR> $\rightarrow$ <D-PART>? saat <NUM>}    \addtag
\]
\[
\texttt{<TIME-EXPR> $\rightarrow$ <DAY>? <D-PART> saat<SUF>*}    \addtag
\]
\[
\texttt{<TIME-EXPR> $\rightarrow$ <DAY>? <D-PART><SUF>*}    \addtag
\]
\[
\texttt{<TIME-EXPR> $\rightarrow$ <DATE-EXPR> <TIME-EXPR>}    \addtag
\]

\begin{table}[!h]
  \caption{Sample \emph{Time} Expressions in Turkish.}
  \label{tab:times}
    \centering
    \begin{tabular}{p{0.21\linewidth}p{0.24\linewidth}p{0.37\linewidth}P{0.07\linewidth}}
    \hline
    \emph{Time Expression}&\emph{Meaning}&\emph{TIMEX3 Annotation}&\emph{Pattern}\\
    \hline
    11.30&11.30&\texttt{<TIMEX3 tid="t1" type="TIME" value="T11:30">11.30</TIMEX3>}&(8)\\
    \hline
    sabah saat dokuz&nine o'clock in the morning&\texttt{<TIMEX3 tid="t1" type="TIME" value="T09:00">sabah saat dokuz</TIMEX3>}&(9)\\
    \hline
    sabah saatleri&morning hours&\texttt{<TIMEX3 tid="t1" type="TIME" value="TMO">sabah saatleri</TIMEX3>}&(10)\\
    \hline
    Pazartesi sabah{\i}&Monday morning&\texttt{<TIMEX3 tid="t1" type="TIME" value="XXXX-WXX-1TMO">Pazartesi sabah{\i}</TIMEX3>}&(11)\\
    \hline
    2 May{\i}s saat 14:00&14:00 o'clock, May 2&\texttt{<TIMEX3 tid="t1" type="TIME" value="XXXX-05-02T14:00">2 May{\i}s saat 14:00</TIMEX3>}&(12)\\
    \hline
    \end{tabular}
\end{table}

\subsection{Set Expressions}
Below provided are common patterns for the extraction of set expressions and sample set expressions conforming to these patterns are listed in Table \ref{tab:sets}.\\
\noindent
\[
\texttt{<SET-EXPR> $\rightarrow$ <DET> (<T-UNIT> | <DAY> | <MON> | <SEAS>)}    \addtag
\]
\[
\texttt{<SET-EXPR> $\rightarrow$ <T-UNIT><SUF> <NUM> <QUANT>?}    \addtag
\]
\[
\texttt{<SET-EXPR> $\rightarrow$ <DET>? <NUM>? <T-UNIT><SUF> <NUM> <QUANT>?}    \addtag
\]

\begin{table}[!h]
  \caption{Sample \emph{Set} Expressions in Turkish.}
  \label{tab:sets}
    \centering
    \begin{tabular}{p{0.24\linewidth}p{0.21\linewidth}p{0.37\linewidth}P{0.07\linewidth}}
    \hline
    \emph{Set Expression}&\emph{Meaning}&\emph{TIMEX3 Annotation}&\emph{Pattern}\\
    \hline
    her ay&every month&\texttt{<TIMEX3 tid="t1" type="SET" value="P1M" quant="EVERY">her ay</TIMEX3>}&(13)\\
    \hline
    her Pazartesi&every Monday&\texttt{<TIMEX3 tid="t1" type="SET" value="XXXX-WXX-1" quant="EVERY">her Pazartesi</TIMEX3>}&(13)\\
    \hline
    haftada iki kez&twice a week&\texttt{<TIMEX3 tid="t1" type="SET" value="P1W" freq="2X">haftada iki kez</TIMEX3>}&(14)\\
    \hline
    her iki g\"unde bir&once every two days&\texttt{<TIMEX3 tid="t1" type="SET" value="P2D" quant="EVERY">iki g\"unde bir</TIMEX3>}&(15)\\
    \hline
    \end{tabular}
\end{table}

\subsection{Duration Expressions}
The two patterns for the extraction of duration expressions in Turkish are given below and three related samples are provided in Table \ref{tab:durations}.\\

\noindent
\[
\texttt{<DURATION-EXPR> $\rightarrow$ <NUM> <T-UNIT>}    \addtag
\]
\[
\texttt{<DURATION-EXPR> $\rightarrow$ <T-UNIT><SUF>*}    \addtag
\]

\begin{table}[!h]
  \caption{Sample \emph{Duration} Expressions in Turkish.}
  \label{tab:durations}
    \centering
    \begin{tabular}{p{0.20\linewidth}p{0.17\linewidth}p{0.45\linewidth}P{0.07\linewidth}}
    \hline
    \emph{Duration Expression}&\emph{Meaning}&\emph{TIMEX3 Annotation}&\emph{Pattern}\\
    \hline
    iki g\"un&two days&\texttt{<TIMEX3 tid="t1" type="DURATION" value="P2D">iki g\"un</TIMEX3>}&(16)\\
    \hline
    sekiz hafta&eight weeks&\texttt{<TIMEX3 tid="t1" type="DURATION" value="P8W">sekiz hafta</TIMEX3>}&(16)\\
    \hline
    y{\i}llar&years&\texttt{<TIMEX3 tid="t1" type="DURATION" value="PXY">y{\i}llar</TIMEX3>}&(17)\\
    \hline
    \end{tabular}
\end{table}

\section{Open Issues}\label{sec:open}
The open issues on temporal expression extraction from Turkish texts include the following:
\begin{itemize}
    \item   The development of temporal expression extraction and normalization systems is an important open issue for Turkish. A convenient system can be achieved by (i) building a rule-based\slash learning system from scratch, or by (ii) extending an already existing and open-source temporal expression extractor, like HeidelTime \cite{HeidelTime} or SUTime \cite{Chang2013SUTime}, to Turkish, or by (iii) extending an already existing Turkish NER system recognizing date and time expressions, like \cite{Kucuk2009}, to make it a full-fledged temporal expression extractor. Deeper examinations of these tools are definitely necessary to assess the feasibility of each option, yet, the second and the third options currently seem less labor-intensive compared to the first one.
    \item   Due to the agglutinative nature of Turkish, the tokens within the temporal expressions can have sequences of suffixes attached, as demonstrated in the proposed patterns given in the previous section. So, a convenient morphological analyzer should be considered for inclusion into the prospective systems.
    \item   In order to train and test the prospective temporal expression extraction and normalization proposals for Turkish, conveniently annotated corpora in Turkish are necessary. To the best of our knowledge, no such resource, in other words, no Turkish Timebank exists currently. Actually, this lack of annotated corpora is an issue even for the more commonly studied problem of NER on Turkish texts. The only study that describes a publicly-available Turkish corpus (of tweets) annotated with the MUC-style basic named entity types (person, location, and organization names, money and percentage expressions, along with date and time expressions) is presented in \cite{Kucuk2014}. This annotated resource can be used as a starting point to build a Turkish Timebank, though it should be noted that no normalization information exists for the annotated date and time expressions in the current form of the resource.
    \item   After the developments to be carried out within the course of the previous two items above, temporal signals (to be annotated with the \texttt{SIGNAL} tag) and events can be included within the scopes of the system proposals to fully comply with the TimeML specifications. Thereby, a full-fledged temporal expression and event extraction system can be achieved for Turkish.
\end{itemize}

\section{Conclusion}\label{sec:conc}
Temporal expression extraction is an important information extraction task and the corresponding extraction tools make significant contributions to larger natural language processing tasks. In this paper, we present a TimeML-based analysis of temporal expressions in Turkish as related studies on Turkish texts are quite rare. We first describe a temporal lexicon and then use the classes in the lexicon as the building blocks to devise a total of 17 wide-coverage patterns for the extraction of date, time, set, and duration expressions in Turkish. We also provide samples of temporal expressions in Turkish along with the related open issues.

\bibliographystyle{lrec2006}

\end{document}